\documentclass{article}
\usepackage{graphicx} % Required for inserting images
\usepackage[sorting=none]{biblatex}
\usepackage{graphicx}
\usepackage{caption}
\usepackage{comment}
\usepackage{multicol}
\usepackage{amsmath}
\usepackage{subcaption}
\usepackage{amssymb}
\usepackage{url}
\usepackage{algorithm}
\usepackage{algpseudocode}
\usepackage{microtype}
\usepackage[top=0.5in, bottom=0.5in, left=0.75in, right=0.75in]{geometry}
\title{Multi-Agent Reinforcement Learning for Autonomous UAV Exploration in Wildfire Response}
\author{Caden Chandra, Jerry Ng}
\date{September 2026}

\begin{document}
\begin{twocolumn}
\maketitle
\begin{abstract}
Wildfires pose escalating challenges for emergency response due to rapid spread, resource limitations, and communication breakdowns. This study develops a deep reinforcement learning (DRL) framework for training unmanned aerial vehicle (UAV) agents to navigate and monitor simulated wildfire environments. We hypothesize that reward shaping and environmental complexity jointly govern UAV convergence behavior, adaptive navigation performance, and the relative influence of individual reward components in multi‑agent wildfire simulations. Using a multi‑agent actor–critic architecture, UAVs share information to improve collective exploration and adapt to dynamic fire conditions. Over training, UAV agents demonstrated stable convergence, with actor loss decreasing from  $10^{-1}$ to $10^{-4}$, critic loss stabilizing near $10^{-2}$, and average reward rising to approximately 250,000. Drones maintained a 100\% success rates and consistent 300‑step episodes while learning structured fire boundary tracking and adaptive repositioning behaviors. Learned policies exhibited consistent navigation patterns such as fire boundary tracking and adaptive repositioning around expanding ignition zones, contributing to improved situational awareness across scenarios. Overall, these findings highlight the potential of multi‑agent DRL as a computational tool for studying UAV coordination in complex wildfire environments. The results suggest that both reward design and environmental structure play critical roles in shaping policy stability and exploration efficiency, offering insights that may inform future research on autonomous wildfire monitoring and emergency response systems.   
\end{abstract}


\section{Introduction}

During the 2025 Los Angeles wildfires, firefighters were hampered by a combination of logistical delays, communication breakdowns, and resource constraints \cite{laedc2025wildfires}. The rapid spread of fire outpaced the ability of crews to respond before fires grew uncontrollably. The slow process of drafting and disseminating operational plans resulted in outdated information that lagged real-time developments \cite{QIU2025100835}. Manpower was further strained as crews were tasked with patrolling for flare-ups, diverting personnel from active firefighting efforts. The intense labor demand led many firefighters to endure 24-hour shifts in hazardous terrain, contributing to severe fatigue. This fatigue impaired reaction times and increased the risk of injury \cite{10.1071/WF24212}. Additionally, essential fire helicopters require frequent maintenance, reducing field availability, and further slowing response times. These compounding issues underscore the urgent need for intelligent, adaptive systems to support real-time decision-making and resource allocation. 

Wildfire environments present a uniquely challenging setting for autonomous systems. The boundaries of fires shift rapidly; wind and terrain introduce uncertainty, and UAVs must balance multiple competing objectives such as safety, exploration, and energy efficiency. Traditional rule-based or static optimization approaches struggle to adapt to these dynamic conditions. Model predictive control (MPC), for example, requires a new solution to be generated for each fire scenario, since no two fires share the same spread patterns or environmental obstacles.  Moreover, once an MPC solution has been formulated, its higher computational cost compared to reinforcement learning (RL) makes it impractical for applications that require real-time updates. By integrating RL into wildfire response, drones gain the ability to adapt in real time, making them more effective for monitoring, early detection, and situational awareness in unpredictable scenarios \cite{RAOUFI2025105076}\cite{REITER2026101045}.  

Drones have increasingly been deployed in wildfire response efforts. Human-piloted UAVs are used to detect hotspots, while others are equipped with tools to suppress flames. However, these systems are limited by their reliance on human operators, which reduces operational uptime and scalability. Recent advances in artificial intelligence (AI), particularly through the integration of unmanned aerial vehicles (UAVs) and deep learning models, have accelerated the development of more autonomous and effective wildfire management strategies \cite{PerformFireFight}. 

Despite these advancements, current unmanned aerial systems (UAS) face limitations in wildfire monitoring due to bandwidth constraints and the continued need for human oversight. To address these challenges, recent initiatives aim to establish hierarchical platforms of multiple UAVs for sustained fire coverage, develop low-computation real-time collaborative learning algorithms for onboard fire detection and mapping, and transmit final fire maps to relevant stakeholders. 

In parallel, modeling efforts have focused on predicting fire spread under extreme conditions using probabilistic surrogate models, which approximate mathematical models since the outcome of interest cannot be easily computed \cite{xu2025generativeaipillarpredicting}\cite{yu2025probabilisticapproachwildfirespread}. To reduce latency in rapidly evolving fire scenarios, large language model (LLM) enhanced agents have been proposed for real-time decision-making in dynamic environments \cite{zheng2025llmenhancedrapidreflexasyncreflectembodied}\cite{Ping2025MultimodalLL}. 

Multimodal LLMs have recently been integrated into UAV swarms to enable intelligent coordination and collaborative observation of wildfires \cite{lankipalle2025swarmfusionrevolutionizingdisasterresponse}. This approach could significantly improve situational awareness and response efficiency. Furthermore, swarm intelligence has shown promise in search-and-rescue operations, offering scalable solutions to locate individuals in disaster zones \cite{yu2025probabilisticapproachwildfirespread}. 

We hypothesize that reward shaping and environmental complexity jointly govern UAV convergence behavior, adaptive navigation performance, and the relative influence of individual reward components in multi‑agent wildfire simulations. To test the hypothesis, we trained multi‑agent DRL UAVs across several simulated fire scenarios—including continuous fire fronts, spotting driven “jump” fires, and dispersed ignition clusters—and compared their resulting navigation behaviors. To further verify this hypothesis, we also tested multiple reward configurations through a systematic ablation study, allowing us to isolate how each reward component influenced UAV convergence, exploration efficiency, and fire‑tracking behavior. Results show that reward formulations emphasizing coverage efficiency and boundary tracking produced more stable learning curves, faster convergence, and higher average rewards. Training produced stable convergence and adaptive navigation behaviors, with UAVs learning to patrol fire perimeters and reposition dynamically as conditions evolved, with average distance to fire decreasing from 7.5 units to 1.0. Moreover, drones learned to maintain safe distances while patrolling fire perimeters, demonstrating adaptive navigation behaviors that improved situational awareness. Overall, the project demonstrates that different wildfire structures require distinct policy and reward designs to deploy UAVs most effectively for autonomous monitoring and exploration. 

\section{Process}
\subsection{Environment}
\subsubsection*{Scenario Design}

\begin{figure*}[htbp]
  \centering
  \begin{subfigure}[t]{0.32\textwidth}
    \centering
    \includegraphics[width=0.9\linewidth]{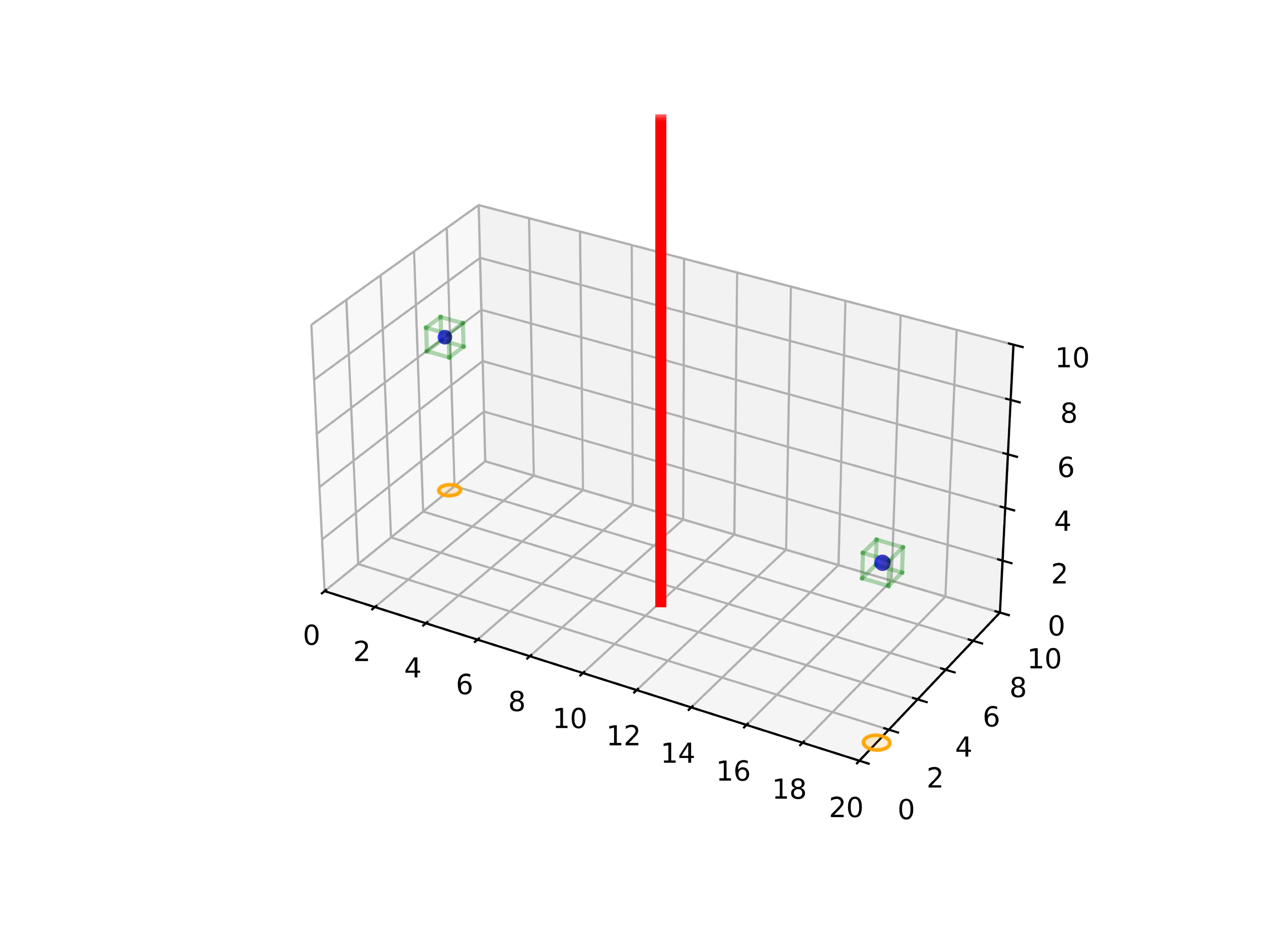}
    \caption{Scenario 1}
  \end{subfigure}\hfill
  \begin{subfigure}[t]{0.32\textwidth}
    \centering
    \includegraphics[width=0.9\linewidth]{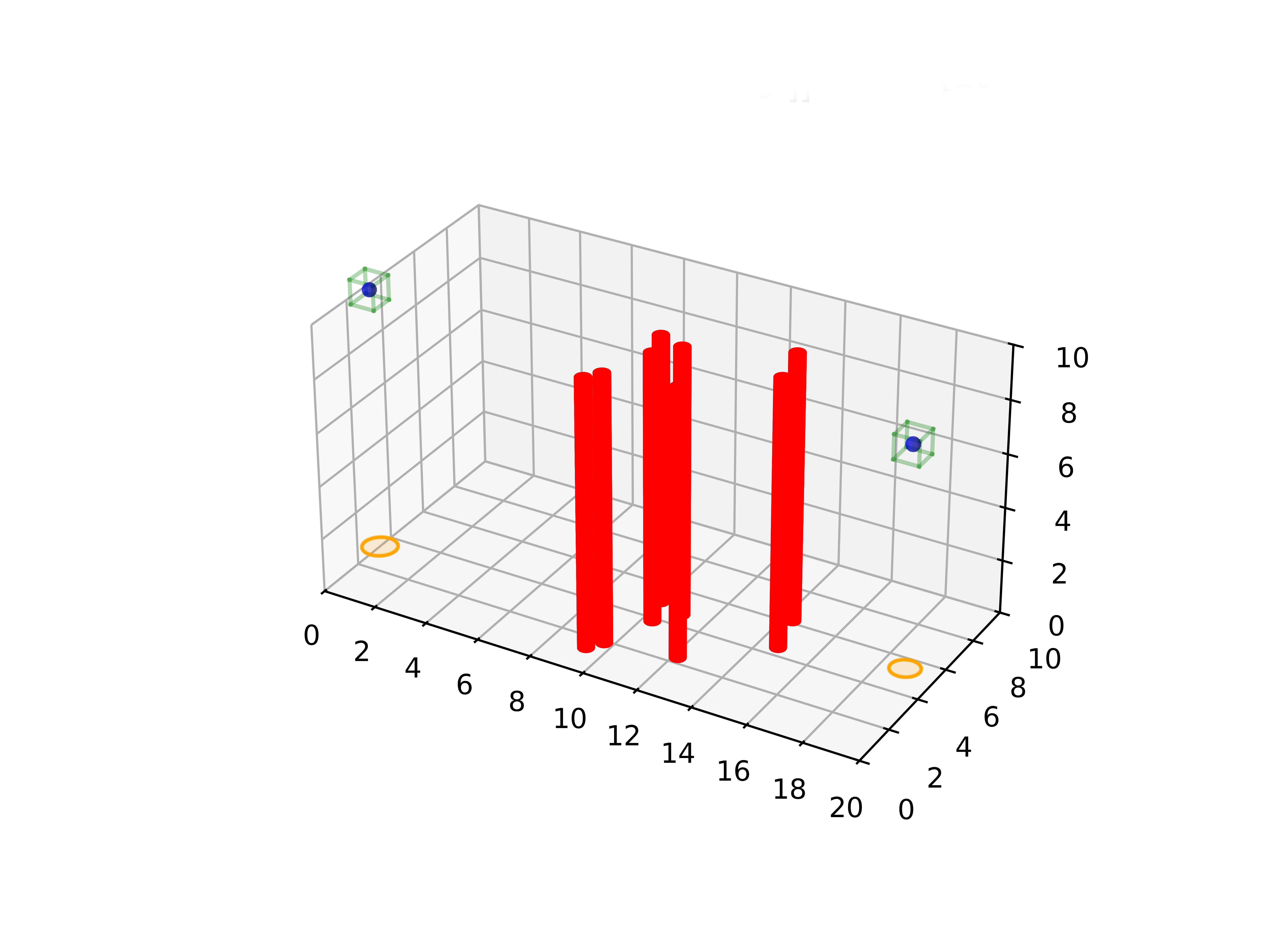}
    \caption{Scenario 2}
  \end{subfigure}\hfill
  \begin{subfigure}[t]{0.32\textwidth}
    \centering
    \includegraphics[width=0.9\linewidth]{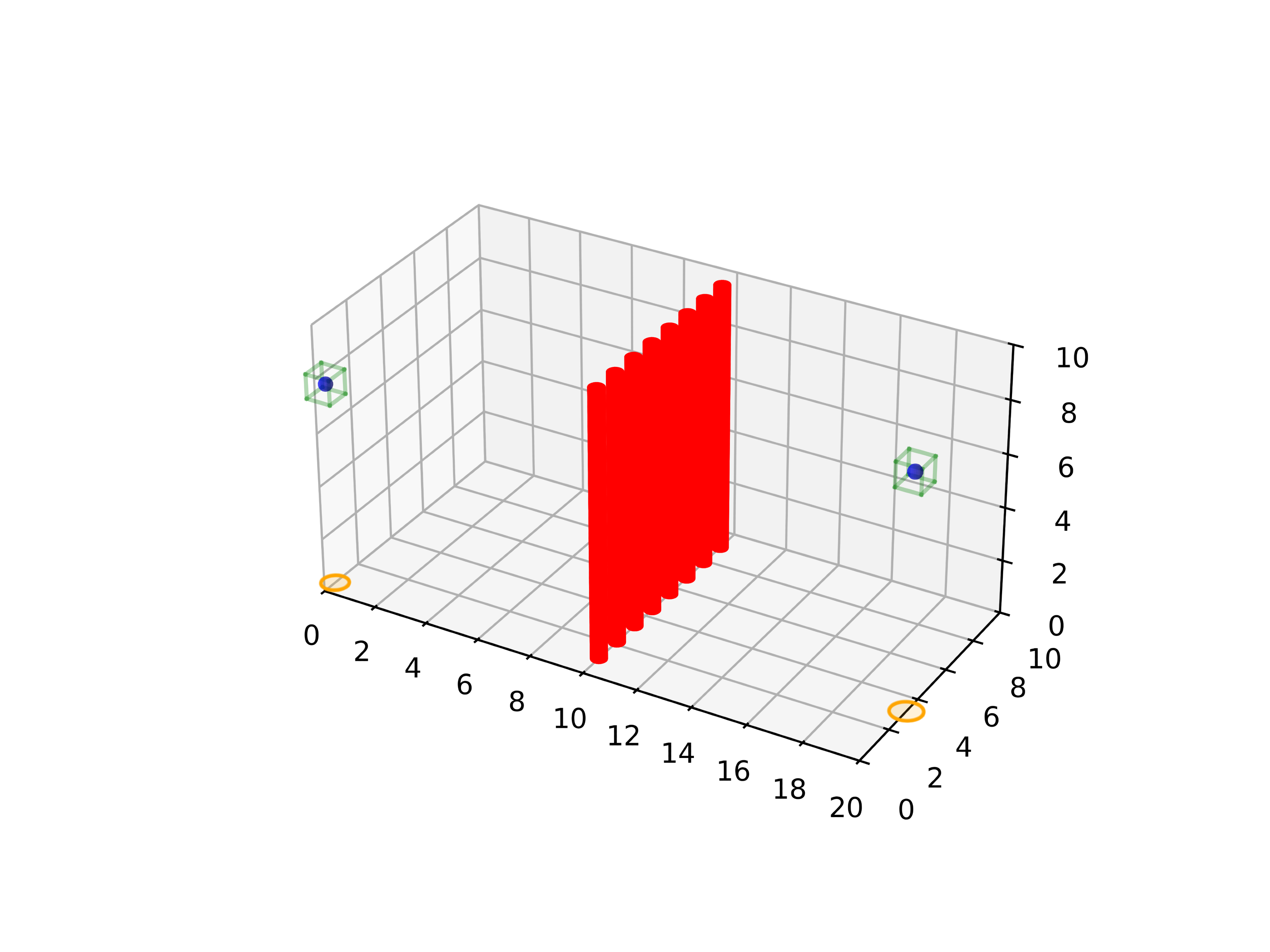}
    \caption{Scenario 3}
  \end{subfigure}
  \begin{subfigure}[t]{0.32\textwidth}
    \centering
    \includegraphics[width=0.9\linewidth]{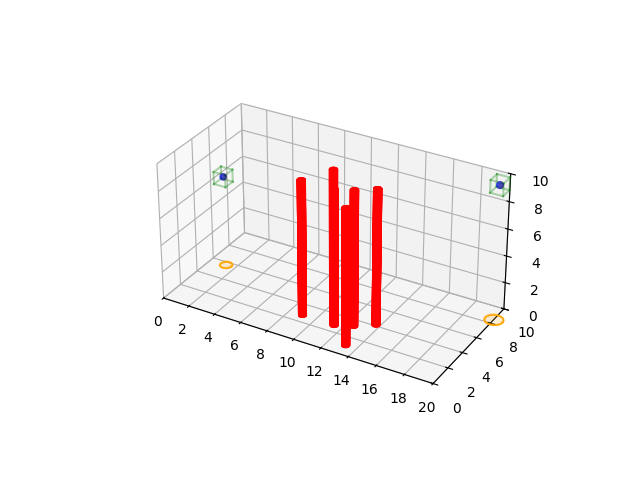}
    \caption{Scenario 4}
  \end{subfigure}
  \begin{subfigure}[t]{0.32\textwidth}
    \centering
    \includegraphics[width=0.9\linewidth]{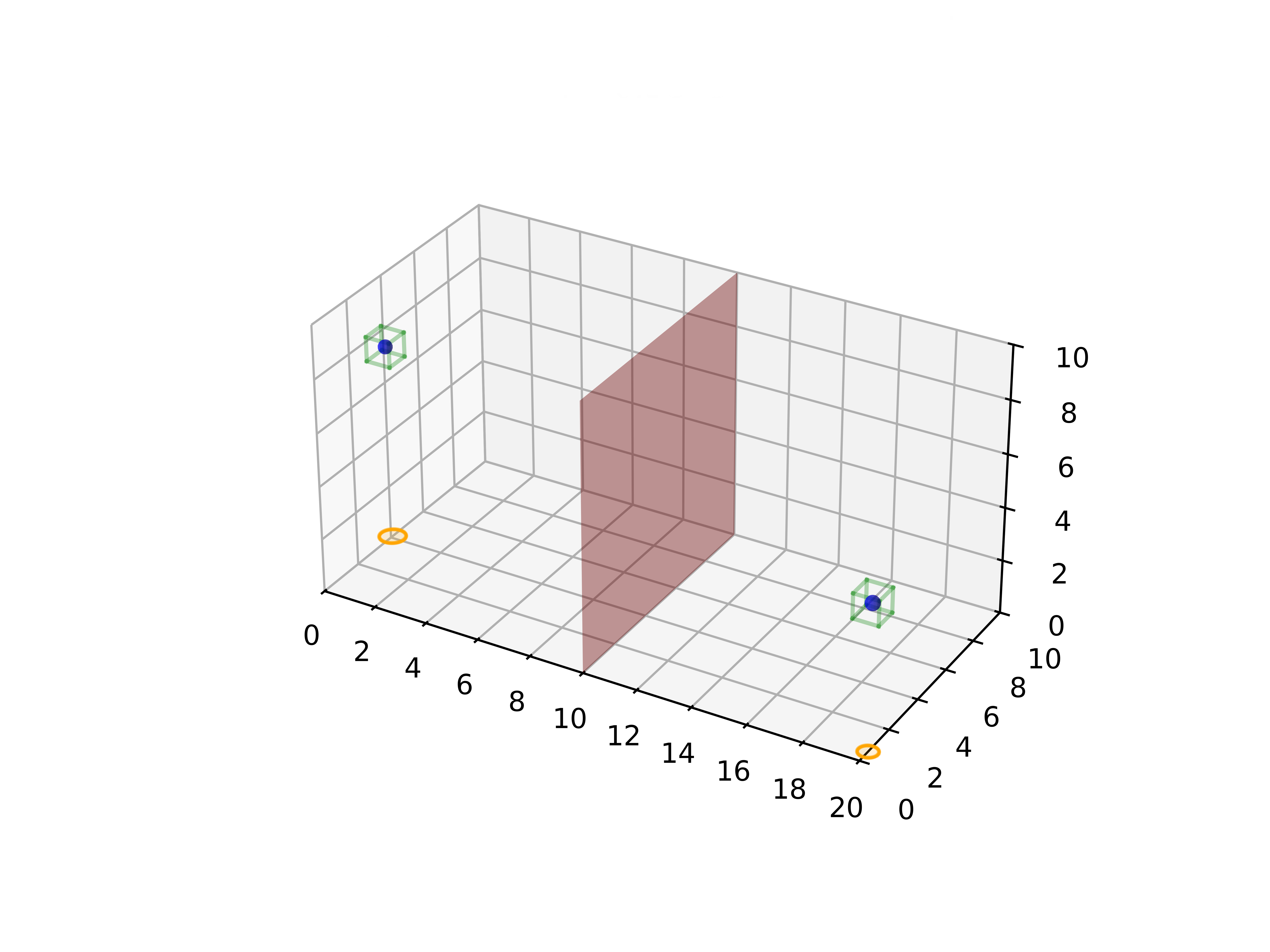}
    \caption{Scenario 5}
  \end{subfigure}
  \begin{subfigure}[t]{0.32\textwidth}
    \centering
    \includegraphics[width=0.9\linewidth]{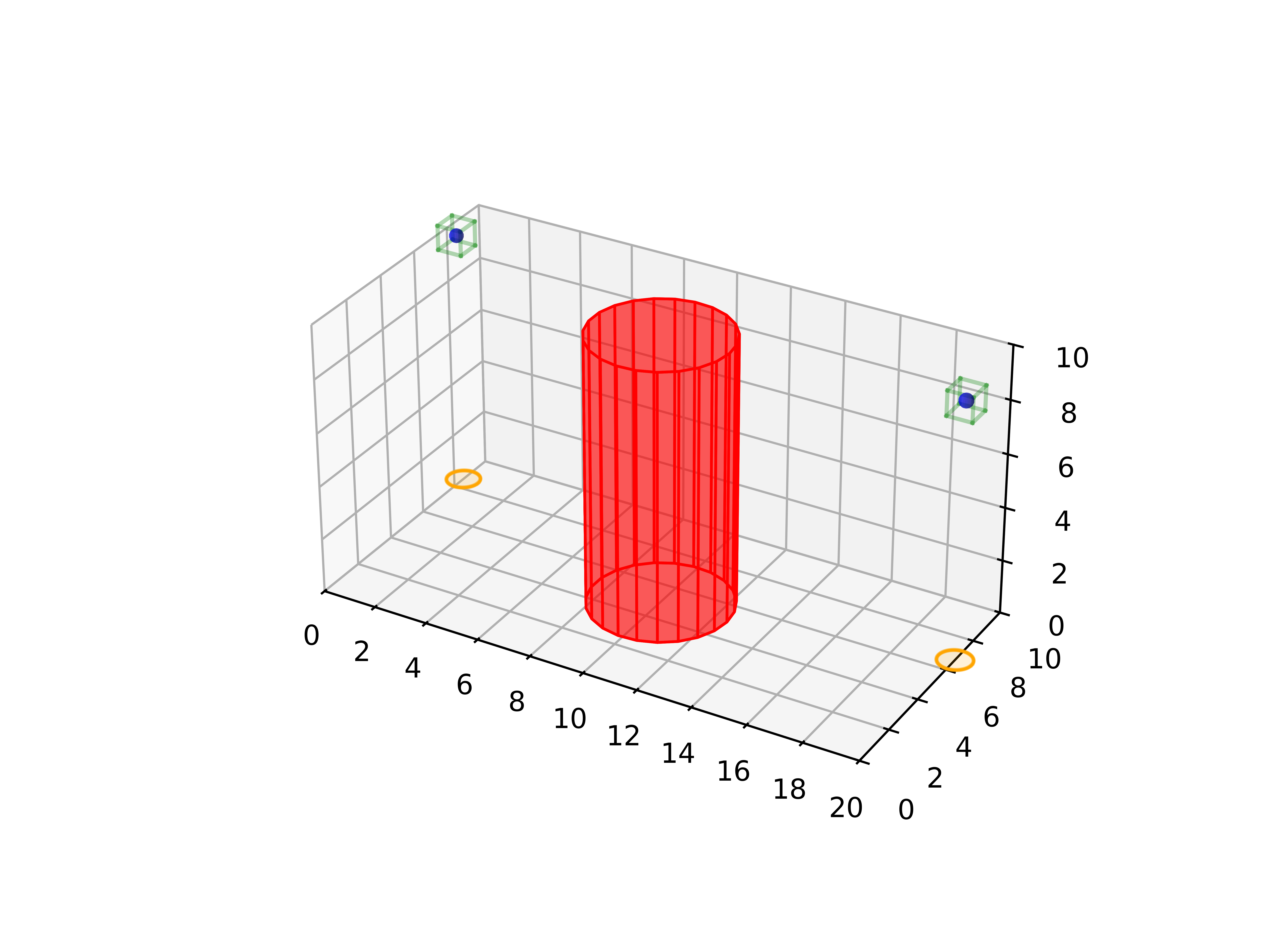}
    \caption{Scenario 6}
  \end{subfigure}
  \caption{Obstacle configuration across six wildfire scenarios, Three dimensional visualizations illustrating the obstacle layouts for scenario 1-6 within the environment. Scenario 1 contains a single central fire; Scenario 2 has randomly spreading fires; Scenario 3 features a wall of 8 fires spaced evenly across the map; Scenario 4 features fires spaced out a distance of 2 units away from the closest fire; Scenario 5 contains an impassable wall of fires; Scenario 6 introduces a central fire with a radius of 2.5. The visualization was captured halfway through the 1000-episode training period (n=1). }
\end{figure*}
Six distinct fire scenarios are implemented:  

Scenario 1: Static Fire — A single fire is placed at the center of the map. The scenario mimics simple fires such as house fires, campfires, or other non-spreading fires.  

Scenario 2: Spreading Fire — A starting fire is located at the center, and new fires randomly spawn within a radius of existing fires in evenly distributed steps. This scenario challenges drones' behavior in unpredictable fires situations. 

Scenario 3: Line of Fires — Eight fires are placed along the x-axis. In this scenario, drones train for when fires are spread in a line.  

Scenario 4: Random Fires — Five fires are randomly placed at the start of each episode. This scenario tests drone exploration and generalization. The scenario mimics a situation where there are fires at many different places, but they are not spreading due to wind.  

Scenario 5: Firewall with Random Crash — An impassable fire wall is placed along the x-axis center. Drones may randomly crash, simulating unpredictable failures. This mimics a scenario where drones are facing a crown fire with many embers spreading to trees and accidently hitting drones 

Scenario 6: Central Fire — A large fire is placed at the center of the map. Drones are trained to circle and patrol around them. This mimics a scenario where fires are growing outwards at a steady pace without wind conditions affecting overall fire spread direction. 

The simulation takes place in a 3D bounded space measuring $20 \times 10 \times 10$ units. Drones are modeled as spheres with a radius of 0.25 units and are assigned bounding boxes in the shape of a cube with a length of 0.5 to simplify collision detection between fires and drones, adopted from Sulaiman et al \cite{6575955}. Fires are represented as vertical cylinders with a default radius of 0.25 units and a height of 10 units. The height is set to the maximum environment size to increase simplicity and mimic the worse case fires which can reach up to 160 feet. In Scenario 6, the fire radius is increased to 2.5, while Scenario 5 introduces a 0.06\% random crash probability per step. These scenarios were selected to represent a range of wildfire behaviors which enables the evaluation of the UAV performance under diverse conditions.  
 
The virtual simulation uses python, gymnasium robotics, numpy, matplotlib, pandas, seaborn, and torch packages. 

At the start of each episode, the environment initializes the fire configuration, drone positions, and reward grid according to the selected scenario. At each timestep, the environment updates fire spread, drone observations, and reward components before returning to the next state. Each drone receives a structured observation vector composed of 10 base entries: its own position 
(x, y, z), a binary indicator for fire presence below, the position of the other drone, and a placeholder fire position. This is extended to up to 12 visible fires, each encoded as 
(x, y, z, visibility flag).

\subsubsection*{Reward Components}

At each timestep, the environment computes rewards based on multiple components: 

Fire-based rewards: Drones are penalized for collisions with fire or entering danger zones near fires and rewarded for maintaining an optimal distance. Collisions result in the immediate termination of the episode. The proximity reward is shaped by a Gaussian function centered at the optimal distance d*: which was adopted and modified from Kim et al \cite{kim2024transformablegaussianrewardfunction}.  

Exploration rewards: The environment is divided into grid cells, each with a regenerating reward value. When a drone visits a cell, its reward is reset to zero and gradually regenerates over time. Drones also receive a discovery bonus for identifying previously unseen fires. 

Edge penalties: To discourage boundary hugging, drones are penalized for staying too close to the edges of the environment. 

Energy penalties: A per-step energy cost is computed based on the change in action magnitude. This penalty is scaled relative to an exponentially moving average baseline to avoid discouraging necessary thrust and was adopted from Brostons et al \cite{moralesbrotons2024exponentialmovingaverageweights}.  

Movement bonus: Drones receive a small bonus for purposeful motion and a mild penalty for remaining stationary. The Euclidean distance is calculated between its current and previous position and then compared against the movement threshold, which specifies the minimum distance required to be considered actively exploring. If the drone moves less than 0.75, a small stationary penalty is applied. Conversely, if the drone moves more than 0.75 units, a positive movement bonus is awarded.  

\textbf{Total reward:} The overall reward $R_t$ at each timestep is computed as a weighted sum of normalized components: 
\begin{equation}
\begin{aligned}
R_t &= w_{\text{prox}} \cdot r_{\text{prox}} + w_{\text{expl}} \cdot r_{\text{expl}} - w_{\text{energy}} \cdot r_{\text{energy}} \\
    &\quad + r_{\text{move}} - r_{\text{edge}} + r_{\text{discover}}
\end{aligned}
\end{equation}
where each term corresponds to the components described above.

To promote exploration and policy diversity, Gaussian noise is added to the base action during candidate generation:
\begin{equation}
    a_i = a_{\text{base}} + \epsilon, \quad \epsilon \sim \mathcal{N}(0, \sigma^2 I)
\end{equation}

where $a_{\text{base}}$ is the action output by the actor network, and $\epsilon$ is sampled from a zero-mean Gaussian distribution. Gaussian noise is added to the actor’s proposed action to generate diverse candidate actions during training. This noise is applied consistently throughout training and is not annealed. The critic then evaluates these candidates, balancing exploration with reward‑driven behavior. 

The curriculum learning strategy gradually reduces the safety margin across five levels, increasing task difficulty and encouraging drones to develop fine-grained control near hazardous regions:
\begin{equation}
   \text{safety\_margin} = \text{base\_margin} + \max(0, (5 - c)) \cdot 0.1 
\end{equation}
where $c$ is the curriculum level. Algorithm 3 demonstrates the process the total process of calculating rewards per step.

In Scenario 2, each existing fire center has a 0.02\% chance of spawning a new fire at each timestep. When a spawn event occurs, a random fire is sampled uniformly within a 5-unit radius from the existing fire center. To prevent cluttering, the new fire is added only if it starts 1.5 units away from all existing fires. 

\subsection{Replay and Normalization}
Each agent stores its experience in a replay buffer, which includes observations, actions, clipped rewards, next obstacle states, and episode termination flags.  

Critic updates follow a Twin Delayed Deep Deterministic Policy Gradient architecture, using clipped target Q-values and two critic networks adopted from Ying et al \cite{ying2025urplanneruniversalparadigmcollisionfree}.  During each episode, Augmented Policy Exploration and Evaluation (APE2) is used to generate multiple action candidates with different Gaussian noise levels. Then it is evaluated by multiple critics. 

The environment uses Welford’s online algorithm to compute the mean and variance of three key reward components: proximity, exploration, and energy. These components are normalized individually and then recombined using weighted coefficients. Proximity weight is 1.5, Exploration weight is 1.2 and Energy weight is $10^{-6}$ respectively. Training begins after a replay warm-up of 20,000 environment steps, with normalization starting after 2,000 collected transitions.  

\subsection{Policy and Candidate Selection}

The actor network generates a base action vector for each drone, which serves as the foundation for candidate selection. To promote diversity, stochastic noise and fire directed heuristics are applied at multiple levels — encouraging exploration when distant from fires and retreat when too close. 

Each candidate is first scored based on its simulated immediate reward, incorporating proximity shaping, optimal bonus logic, and penalties for unsafe distances or stationary behavior. 

To evaluate long-term value, each candidate is passed through the critic network, which averages Q-values across multiple critics. A small energy penalty is subtracted from each Q-value to discourage inefficient or erratic movement. Actor-side regularization is also applied to promote smooth control. 

The candidate with the highest adjusted score is selected for execution. Candidate scores were computed using critic estimates and energy penalties, enabling drones to learn robust policies for fire tracking, exploration, and collision avoidance.  

\section{Results} 

The study examined how multi-agent reinforcement learning can optimize drone navigation for wildfire tracking and response. Agents were trained over 5 curriculum levels, with each level having 200 episodes. Figures 2–5 demonstrate that the proposed multi-agent reinforcement learning framework is successful in learning effective wildfire exploration policies. 

\begin{minipage}{0.45\linewidth}
  \centering
  \includegraphics[width=\linewidth]{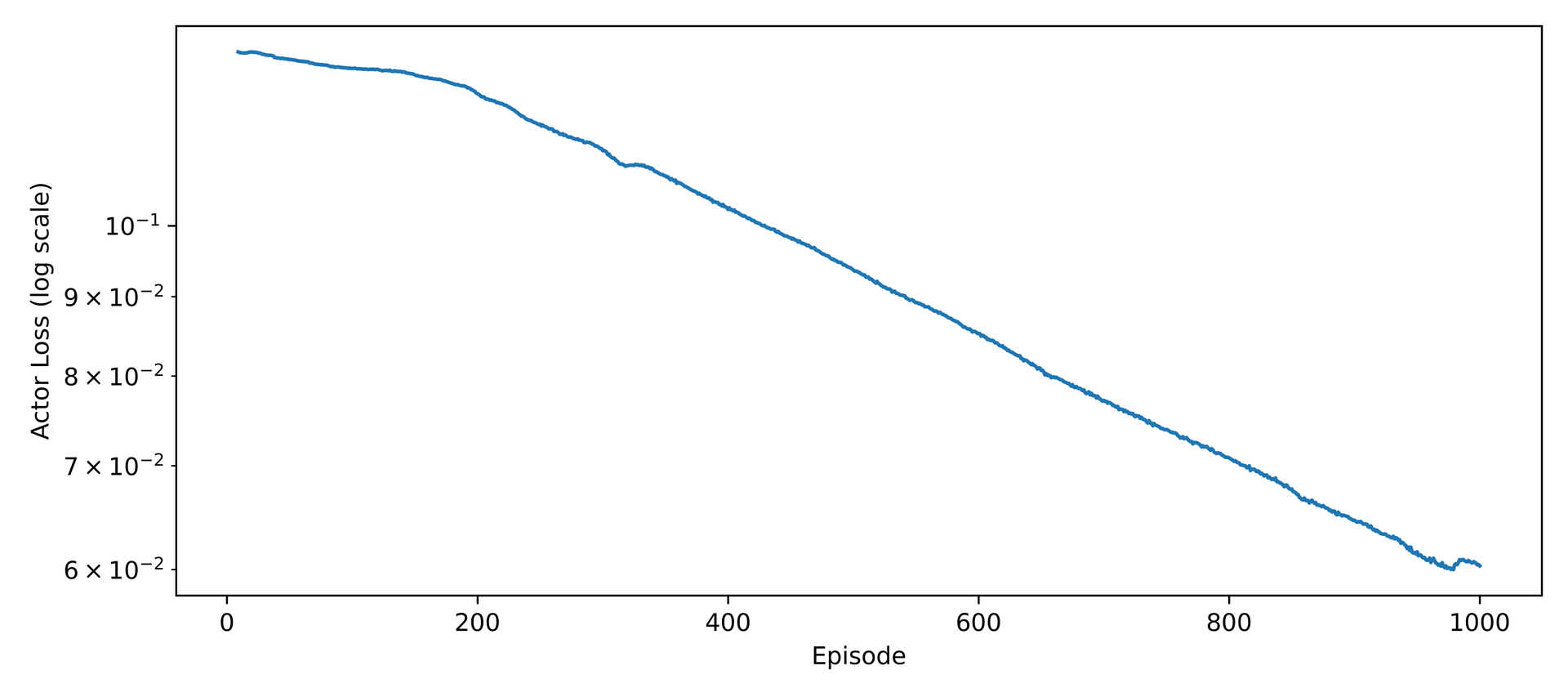}
  \text{(a) Actor loss}
\end{minipage}\hfill
\begin{minipage}{0.45\linewidth}
  \centering
  \includegraphics[width=\linewidth]{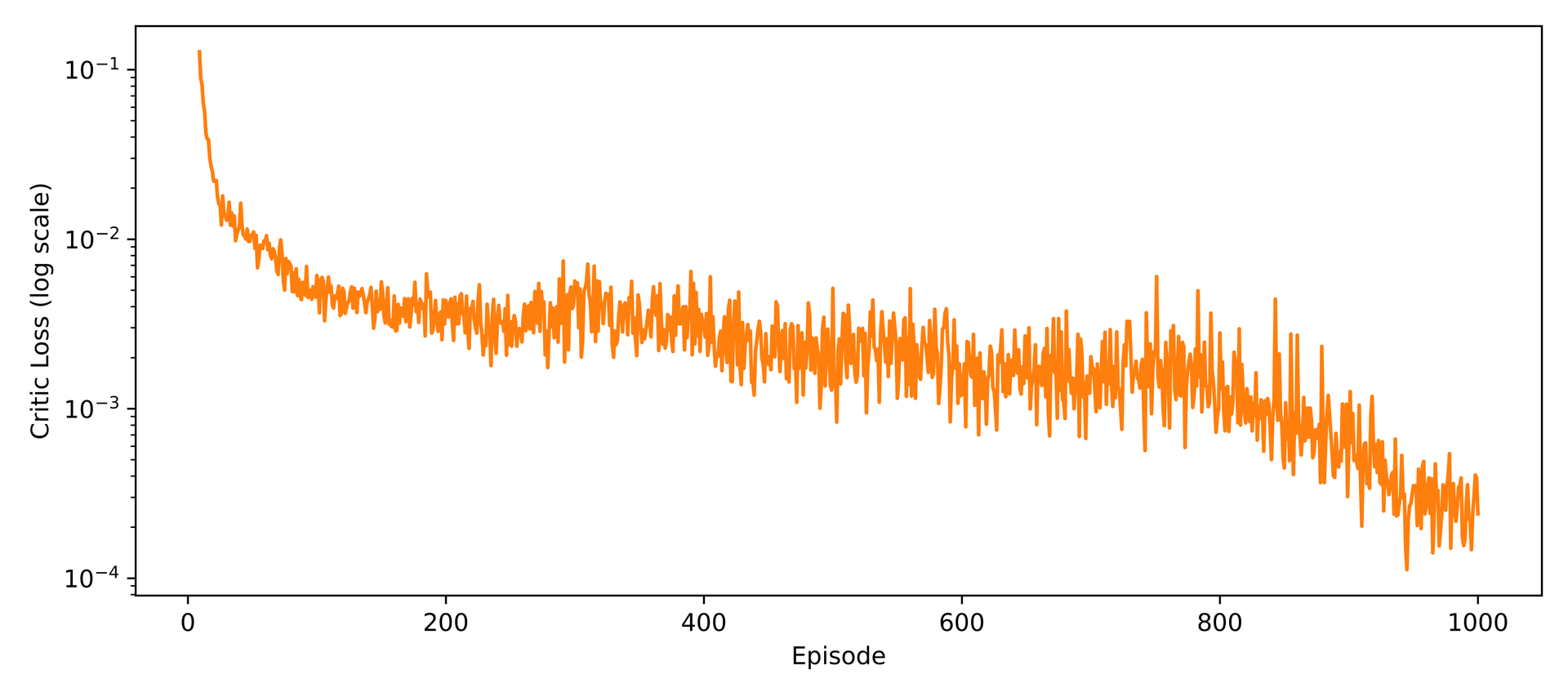}
  \text{(b) Critic loss}
\end{minipage}

\bigskip
\noindent
Figure 2: Actor-critic loss dynamic during training. (a) Critic loss during training episodes plotted on a logarithmic scale. (b) Actor loss during training episodes plotted on a logarithmic scale. Over 1000 training episodes, actor and critic loss were computed through their respective formulas plotted for every episode (n=1).
\bigskip

Figure 2 represents the actor and critic loss over 1000 training epochs. Actor loss holds around $10^{-2}$ after an initial decay from $10^{-1}$ then decays to a final value near $10^{-4}$. Meanwhile, critic loss decreases from $10^{-1}$ and stabilizes near $6 \times10^{-2}$.  

\begin{center}
    \includegraphics[width=\linewidth]{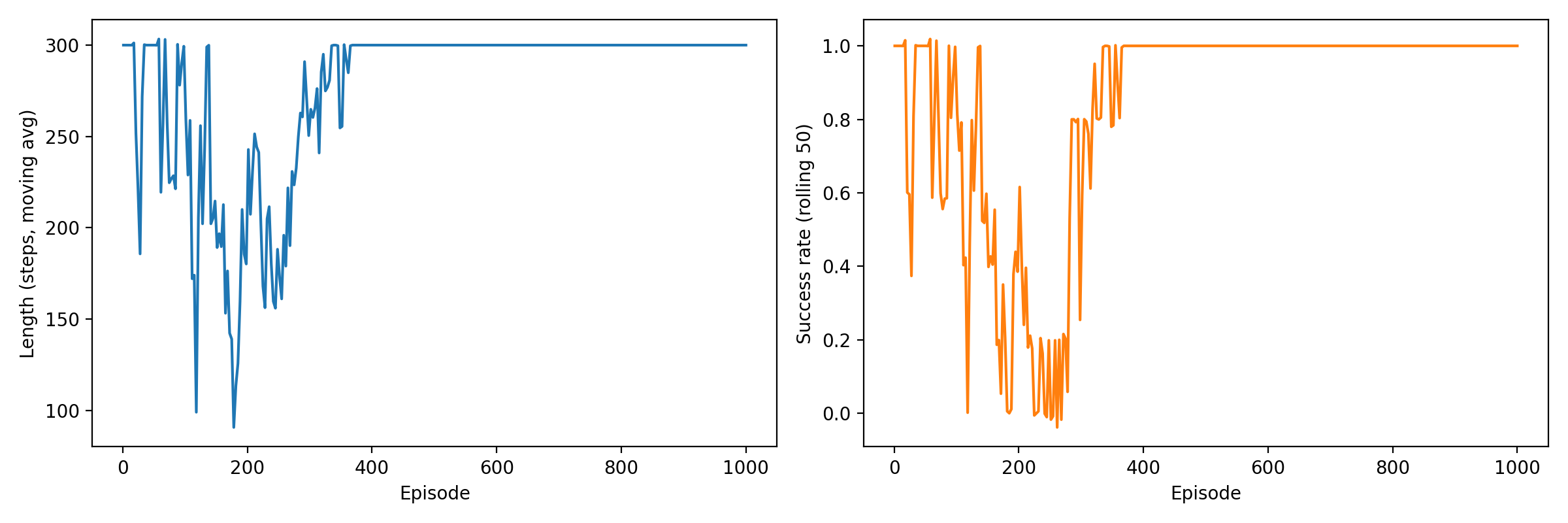}
    \text{(a) length  \hspace{7em} (b) success}
\end{center}
\begin{center}
    \includegraphics[width=\linewidth]{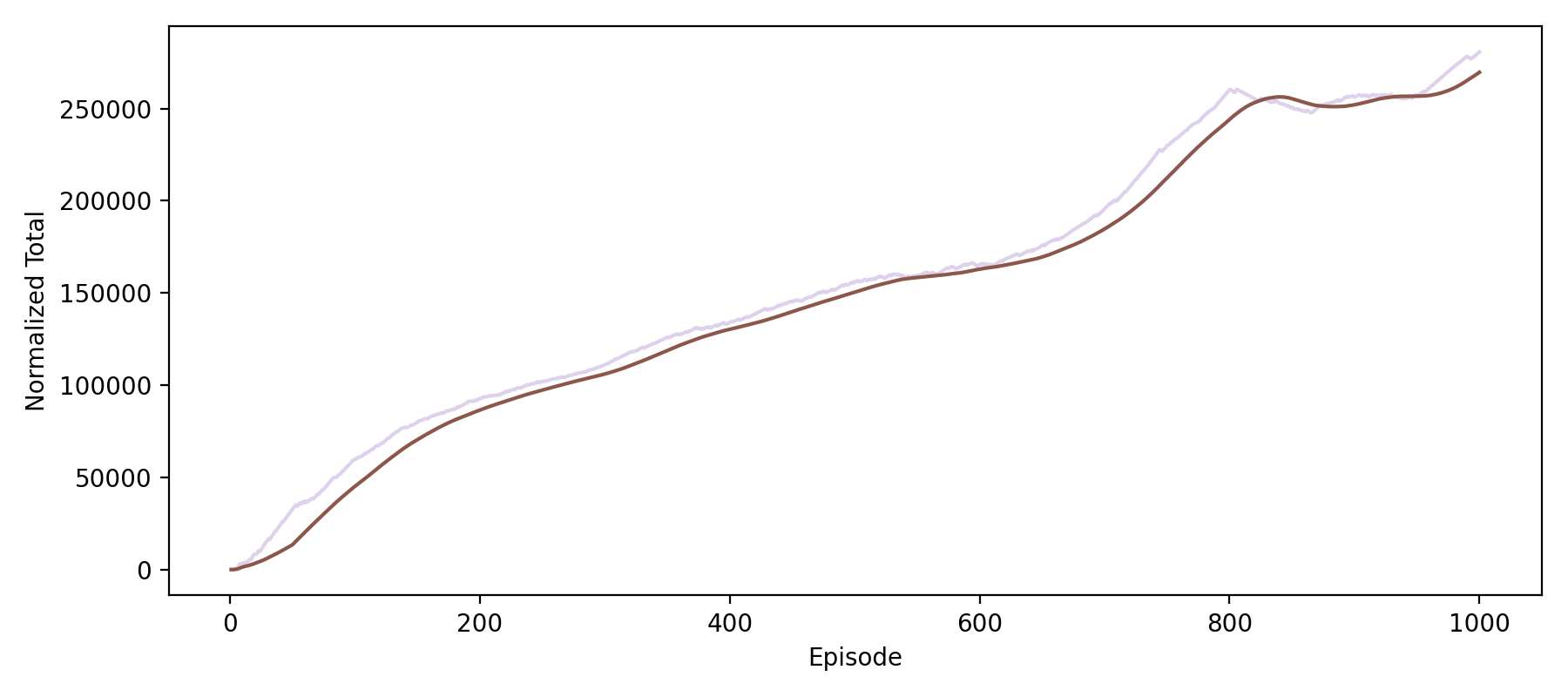}
    \text{(c) reward}
\end{center}
\noindent
Figure 3. Training dynamics of the reinforcement learning agent over 1000 episodes. (a) Episode lengths across training, showing variability in how long the agent remained active with a 20 moving average. (b) Success rate over training period. Success is considered when the drone goes 300 steps without crashing into any object and detects a fire. Success is calculated over a period of 20 episodes. (c) reward through the course of training. All curves were generated from a single training run consisting of 1000 learning episodes (n=1).
\bigskip

Figure 3a, episode length, starts near 300 steps but quickly fluctuates between 70 and 300 steps completed, with an overall declining trend. Starting at episode 240, the length increases from its relative minimum. After 400 episodes, the drone consistently achieves 300 steps through the completion of training. Figure 3b, the success rate, starts at 100\%, but then fluctuates between 0\% and 100\%, showing an overall declining trend. Near episode 200, the success rate fluctuates between 0\% and 20\% but starts increasing after step 250. The success rate stabilized near 100\% at episode 400 and remained constant through the end of the training period. Figure 3c, reward, starts at 0, grows relatively rapidly until episode 100, where the line starts to flatten out. The reward spikes between episode 700 to 800 but stabilizes near 250,000 at episode 800 until the end of training. 
 
\begin{center}
    \includegraphics[width=\linewidth]{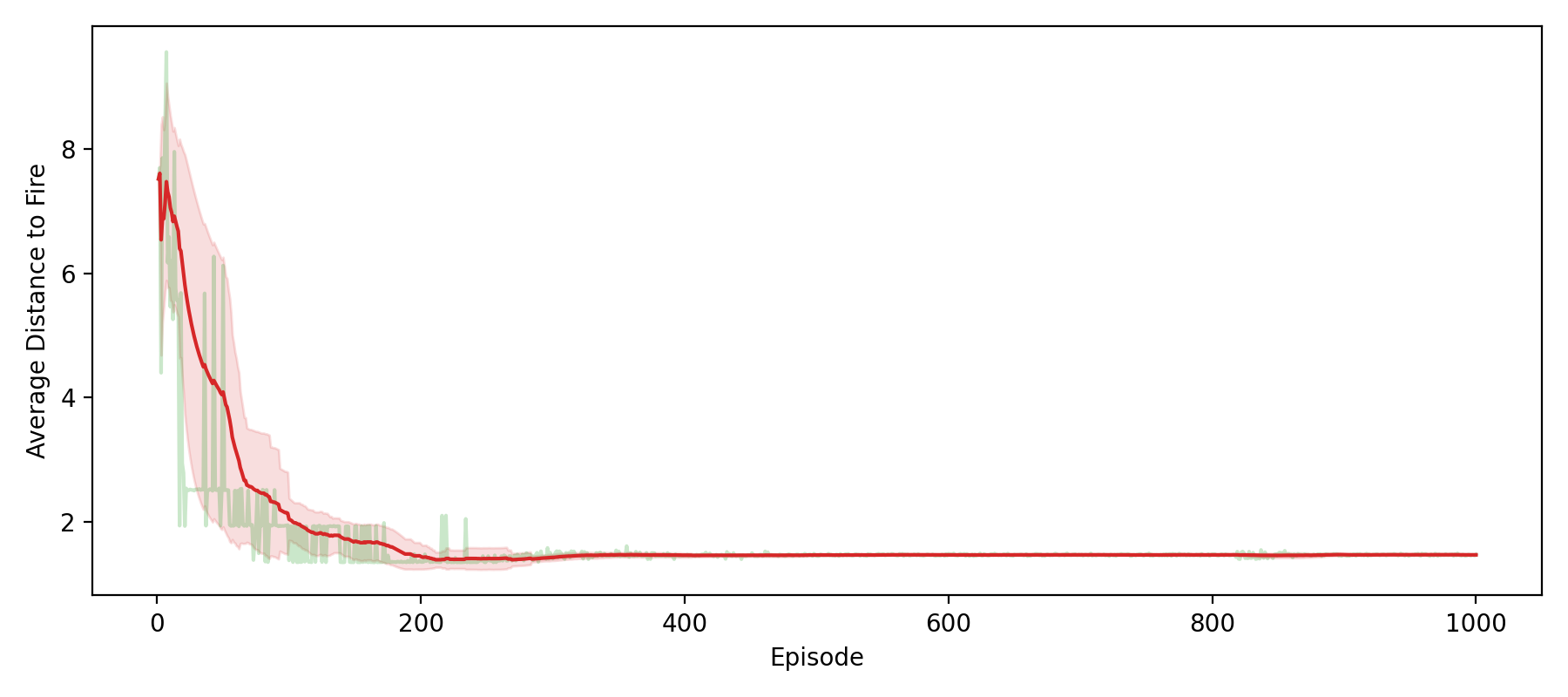}
\end{center}
{Figure 4: Average distance between the agent and active fires during training. Line plot showing the agent’s distance to the nearest active fire across 1000 training episodes. The red curve represents the smoothed mean distance while the shaded region represents the within-episode variability, and the green vertical markers show individual episode-level values. Data was generated from a single training run consisting of 1000 learning episodes (n=1).}
\bigskip

In Figure 4, drones began with an average distance to the fire of 7.5 units. Over the first 200 episodes, the average distance dropped rapidly. During this period individual episodes average distance to the fire varied with inconstant spikes in distance. After episode 200, the distance to the fire began to stabilize at 1.5 with continued, smaller fluctuations. Finally, after 400 episodes, episode variability and within-episode variability converged to 1 with little spread between episodes. 

\begin{center}
    \includegraphics[width=\linewidth]{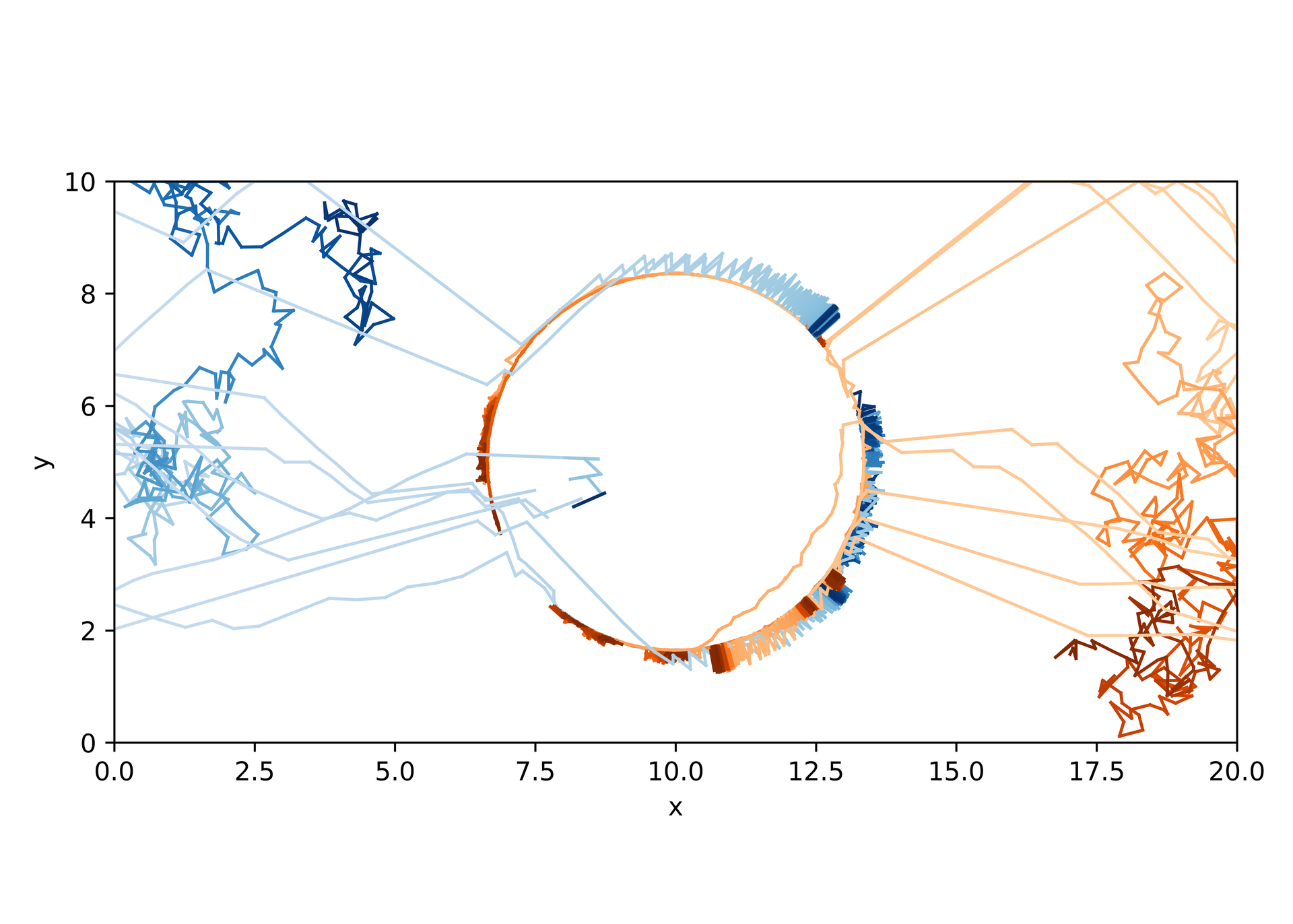}
\end{center}
Figure 5. Drone trajectory around central fire in scenario 6. Trajectories of agent navigating a 20x10 grid throughout training period. The data sources from scenario 6, central fire.  Blue path represents agents initialized on the left side while the red path represents agents initialized on the right. Darker colors represent later episode performance. Trajectories were sampled every 100 episodes from the training of a single policy every 1000 episodes (n=1).
\bigskip

Figure 5 illustrates representative drone trajectories sampled from different stages of training. Early trajectories were short and irregular, often ending in boundary collisions or looping paths far from the fire. As training progressed, trajectories became longer and more structured. Later paths showed more consistent movement patterns, including extended arcs and repeated passes around the fire perimeter.  

The implementation of curriculum levels shifted the trajectory of the drones from hugging the walls throughout the entire training period to gradually exploring closer to the fire. The curriculum levels also reduced the frequency of early training crashes. To further understand how individual reward components shaped these behaviors, we conducted an ablation study. 

\subsection{Ablation Study}
To assess the relative contribution of each reward signal, we performed an ablation study in which the baseline model included all ten components. Each component was then removed individually, and the resulting performance was compared against the baseline. Further tuning of reward weights was not implemented after the ablation study. Instead, we relied on the ablation study to evaluate how each reward component influenced agent behavior. By systematically removing individual reward terms and comparing performance against the baseline, we assessed the relative contribution of each component without modifying the underlying weight structure. The components considered were: Proximity Reward and Penalty (Proximity), which regulates safe monitoring distances; Exploration Cell Rewards (Exploration), which incentivize coverage of new areas;  Per-step Energy Penalty (Energy), which discourages excessive energy use; Edge and Wall Penalty (Edge), which discourages hugging boundaries; Movement Bonus (Movement), which promotes active navigation; Stationary Penalty (Stationary), which discourages immobility; Fire Discovery Bonus (Discovery), which rewards identification of new fires; Lateral Patrol Bonus (Lateral), which encourages perimeter patrolling; Optimal Distance Bonus (Optimal), which reinforces maintaining an ideal monitoring range; and Cell Reward Regeneration (Regeneration), which replenishes exploration incentives to encourage repeated visits over time. Further description of the reward is given in the methodology section. For brevity, we refer to each reward component by its abbreviated designation in the following analysis. 

\begin{figure*}[htbp]
  \centering
  \begin{subfigure}[t]{0.32\textwidth}
    \centering
    \includegraphics[width=0.9\linewidth]{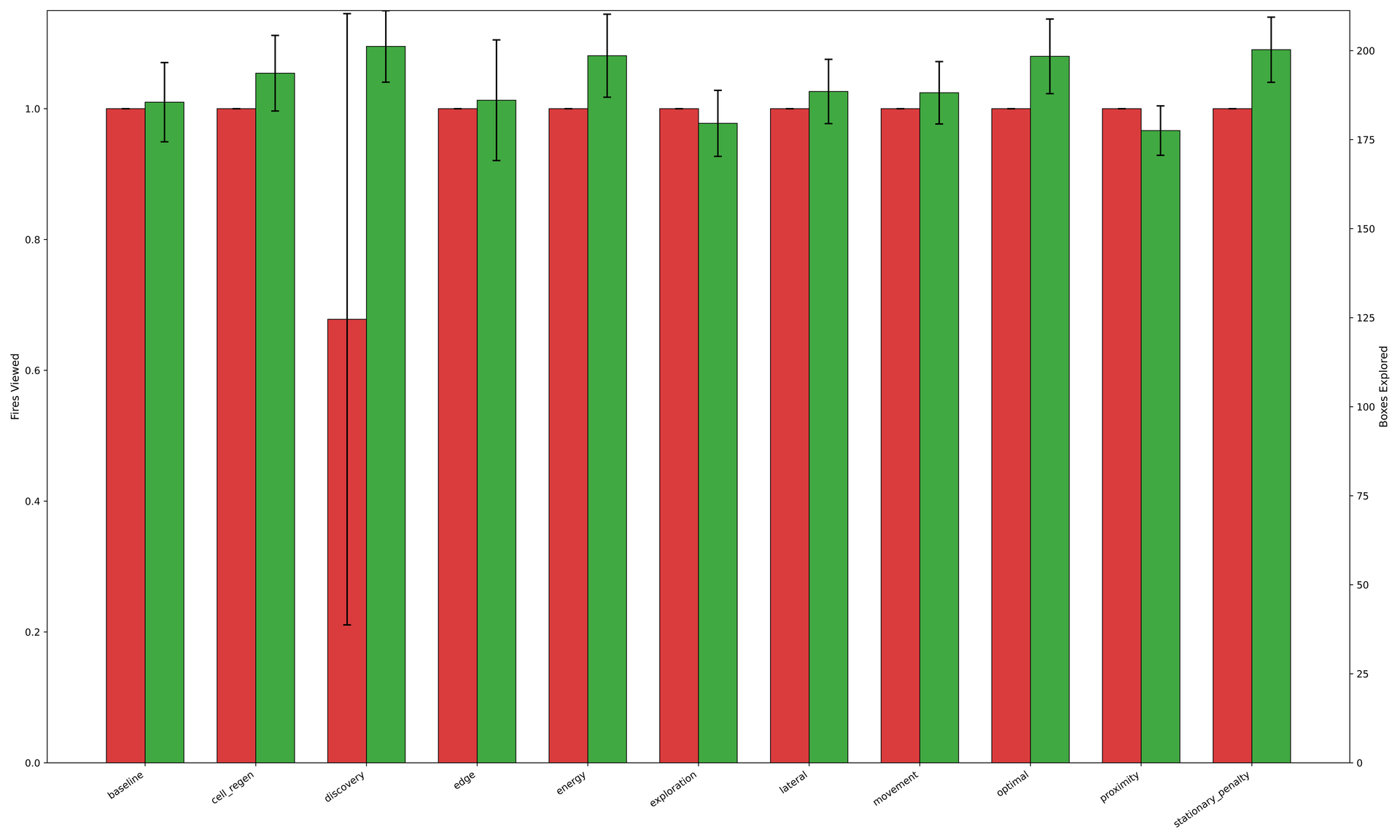}
    \caption{Scenario 1}
  \end{subfigure}\hfill
  \begin{subfigure}[t]{0.32\textwidth}
    \centering
    \includegraphics[width=0.9\linewidth]{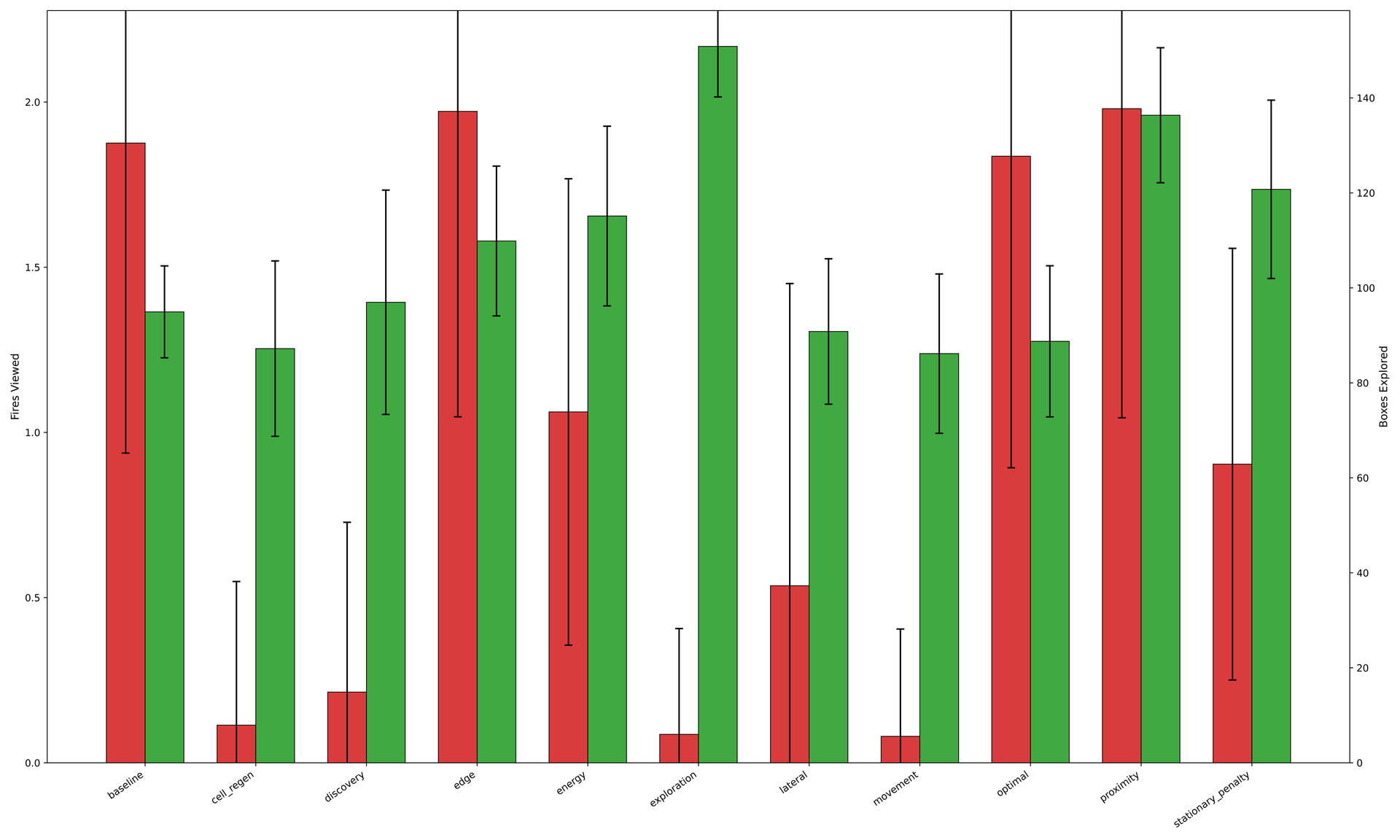}
    \caption{Scenario 2}
  \end{subfigure}\hfill
  \begin{subfigure}[t]{0.32\textwidth}
    \centering
    \includegraphics[width=0.9\linewidth]{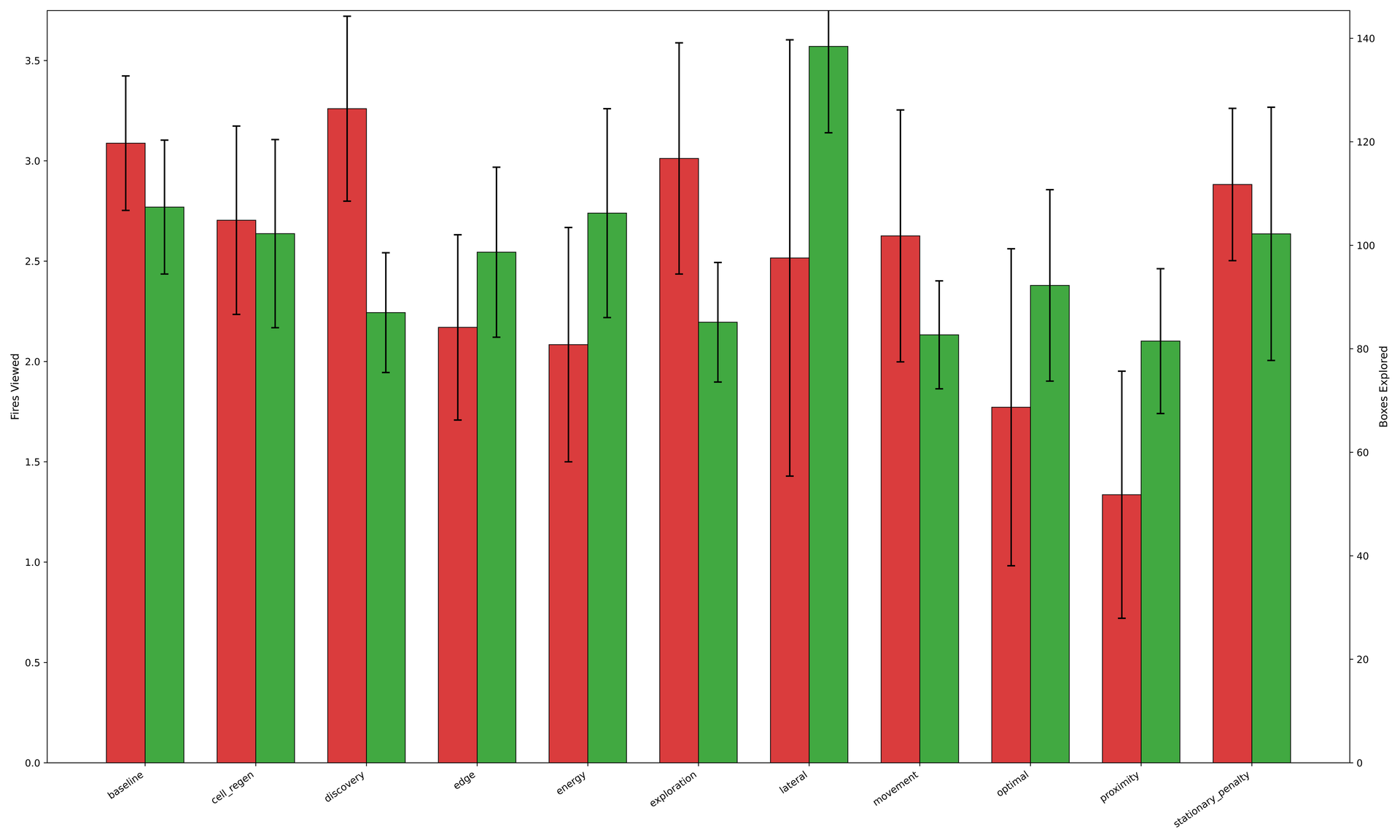}
    \caption{Scenario 3}
  \end{subfigure}
  \begin{subfigure}[t]{0.32\textwidth}
    \centering
    \includegraphics[width=0.9\linewidth]{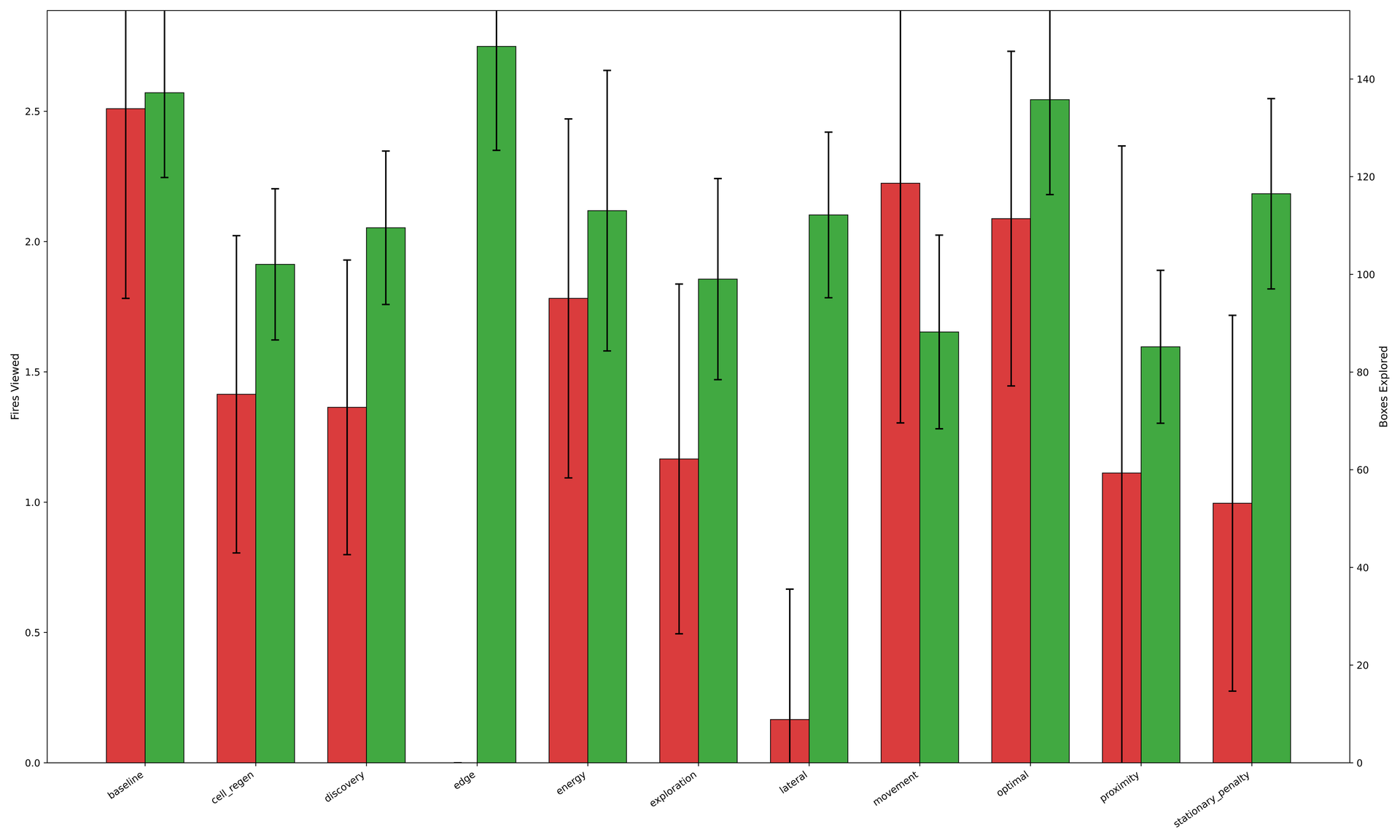}
    \caption{Scenario 4}
  \end{subfigure}
  \begin{subfigure}[t]{0.32\textwidth}
    \centering
    \includegraphics[width=0.9\linewidth]{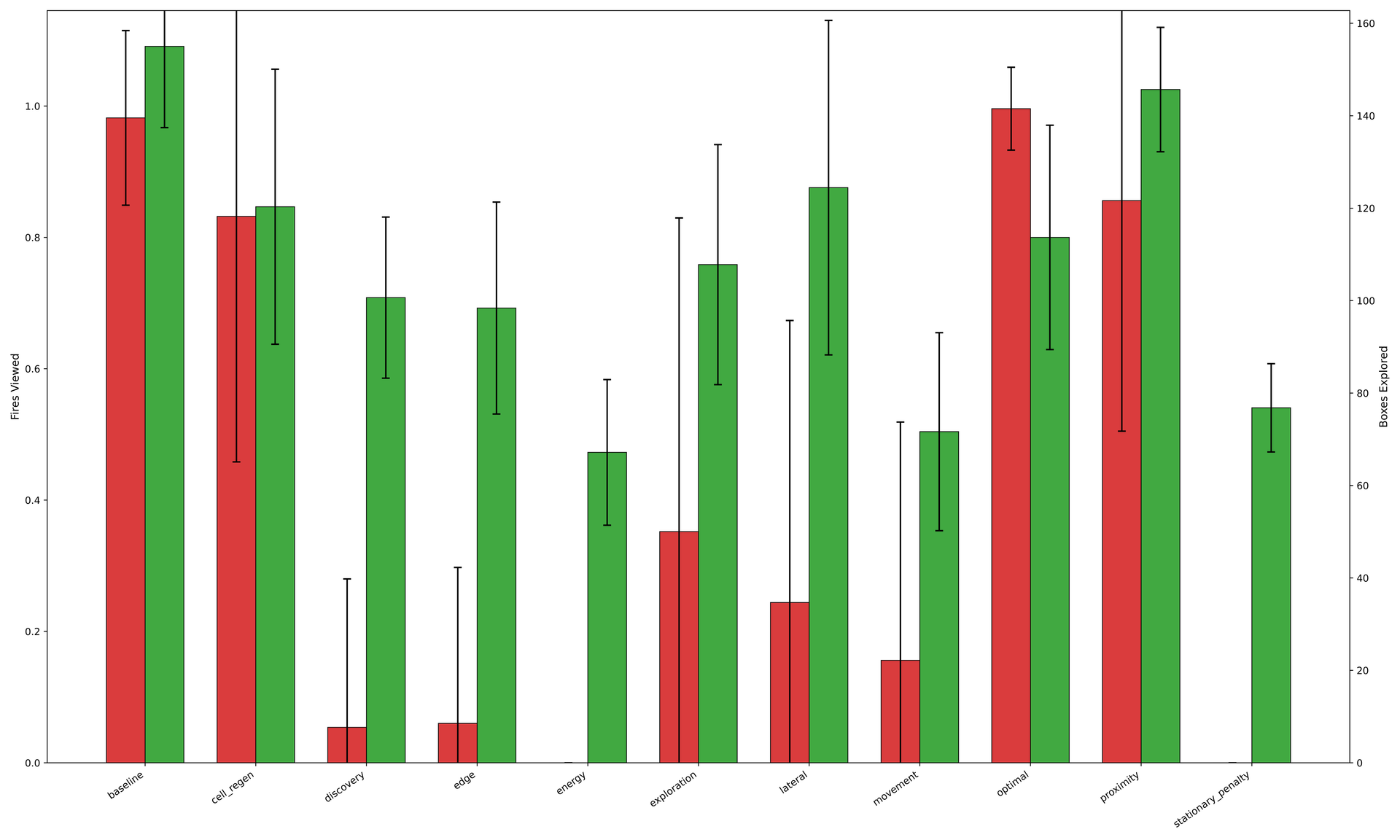}
    \caption{Scenario 5}
  \end{subfigure}
  \begin{subfigure}[t]{0.32\textwidth}
    \centering
    \includegraphics[width=0.9\linewidth]{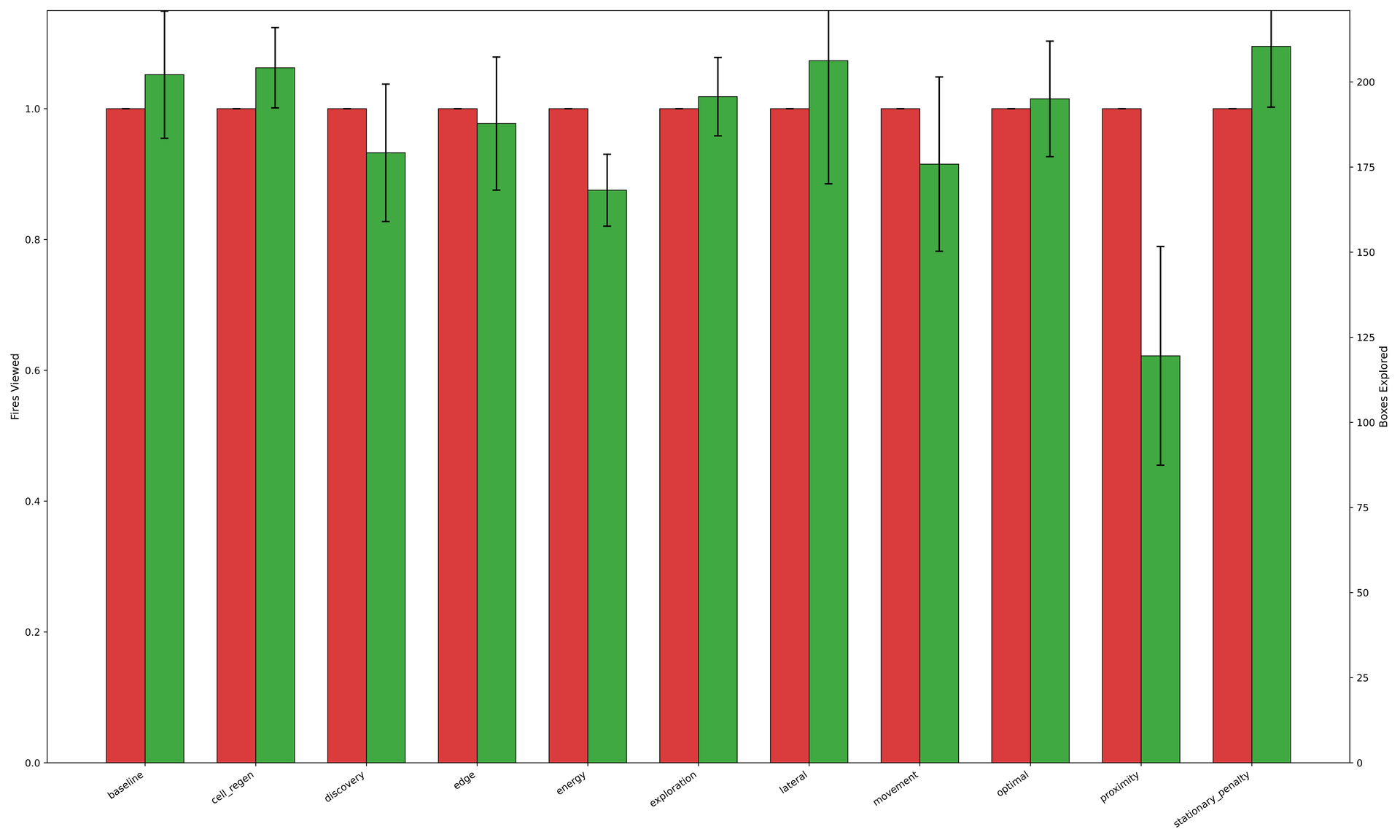}
    \caption{Scenario 6}
  \end{subfigure}
  \caption{ablation study of fire occurrences and box visibility across six wildfire scenarios. Grouped bar chart showing mean ± SD values for fires detected (red) and boxes viewed (green) across scenarios 1-6 using 11 different categories. Each scenario was evaluated using a single trained policy test over 500 deterministic episodes, with bars representing the average value and error bars indicating the variability (n=1).}
\end{figure*}

In Figure 6a, the number of fires viewed remained approximately constant at 1 across all reward categories, with minimal variation except in the Discovery category, which averaged 0.6. For squares visited, most categories clustered near 180, while Discovery again acted as an outlier with the highest mean of 200. In Figure 6b, Baseline, Edge, Proximity, and Optimal achieved the strongest performance, each averaging about 1.8 fires per episode. Exploration exhibited the broadest coverage, viewing 150 tiles—surpassing Baseline by 60 and Proximity by 20. In Figure 6c, Baseline, Discovery, Exploration, and Stationary Penalty produced comparable results of 3.1 fires, with Discovery leading slightly at 3.3. For tiles viewed, Lateral demonstrated the highest average of 140, exceeding the 110 observed under Baseline. In Figure 6d, Baseline again led in fires viewed, 2.5, followed by the next-best movement strategy at 2.2. Edge achieved the greatest spatial coverage, averaging 150 tiles compared to Baseline’s 135. In Figure 6e, Optimal and Baseline performed similarly, each near 0.96 fires—the highest among reward categories. Baseline also viewed the most squares, 150, with Proximity close behind at 143. Finally, in Figure 6f, all reward categories viewed the same number of fires, 1.0, with no deviation. Stationary penalty led in tile coverage with 220, followed by Baseline, Cell Regen, and Lateral by approximately 20 tiles. 

\section{Discussion}

Across all scenarios, training led to increased episode rewards, longer episode durations, reduced boundary collisions, and more frequent movement. Loss curves stabilized, and all scenario-specific metrics showed measurable improvements in drone behavior over the 1000‑episode training period.   

The actor and critic loss curves show that both the policy and value networks converged in a stable manner during training. Fluctuations are typical in multi‑agent systems and reflect normal adaptation rather than instability. The critic loss shows greater variation because its value estimates depend on the actor’s evolving policy with each update shifting the returns it must learn to predict. As training progresses, the decreasing amplitude of these oscillations indicates that the critic is adjusting effectively and that the overall learning process is stabilizing. Episode length and success rate trends show that agents increasingly avoid collisions and maintain exploration without premature termination, while also improving their ability to map and monitor the environment. Figure 3 shows how episode length and success rate evolved throughout training. As training progressed, episodes became longer, indicating that agents were able to remain active in the environment for more timesteps. At the same time, the success rate increased, meaning a larger proportion of episodes ended with the agent achieving the defined goal conditions. Together, these trends suggest that the agent not only survived longer but also became more effective at completing the task.  Longer episodes reflect greater stability and resilience, and higher success rates show that exploration became structured rather than random. The early drop in episode length and success rate is expected in multi‑agent RL, reflecting the drone’s initial exploration and the learning of the fire‑collision penalty; as training progresses, these metrics rise again, indicating stable convergence and improved task performance. Together, these metrics demonstrate that reward shaping and multi‑agent coordination support the development of behaviors that balance safety with environmental coverage. The increasing reward function demonstrates that the agent has learned to avoid colliding with fires while staying within viewing distance and to use a patrolling strategy to discover new fires. The curriculum levels allowed the agent to develop fine-grained control near hazardous regions. Across all metrics, the agents consistently learned to explore, adapt, and patrol wildfire environments, achieving stable convergence, improved coverage, and safety‑aware navigation. 

The ablation study provides insight into how individual reward components shape UAV learning behavior and exploration efficiency. By systematically removing each signal from the baseline model, we observed distinct shifts in performance that revealed which reward components most strongly shaped navigation strategies and policy stability, and which had more limited influence. Collectively, these results highlight how reward composition directly influences navigation strategy—agents rewarded for spatial diversity explore more extensively, while those guided by proximity or balanced objectives detect fires more reliably. Overall, the baseline performed near the top across all six scenarios in both fires viewed and tiles visited, suggesting that the reward weights were properly balanced across diverse conditions. While future work could explore policy-switching strategies that tailor reward categories to specific fire structures, maintaining a unified policy, reducing complexity, and mitigating risks of misclassifications. This phenomenon suggests that switching among multiple policies may require the agent to classify the fire scenario in real time, which can introduce additional uncertainty. Misclassification would cause the drone to apply an inappropriate policy, potentially leading to unsafe navigation or reduced coverage. A unified policy avoids this failure mode and ensures a more consistent behavior across all scenarios. These findings affirm that multi-component reward design supports robust, generalizable UAV behavior across varied wildfire environments.  

\section{Conclusion}

The combination of reward shaping, curriculum progression, and coordinated multi‑agent training enabled the drones to learn reliable strategies for detecting and avoiding wildfires. Several factors likely contributed to these outcomes. The reward function placed strong emphasis on maintaining appropriate distance from the fire, which may have encouraged boundary‑patrolling behaviors over alternative strategies. The deterministic fire‑spread model also simplified learning by providing predictable patterns for agents to track. Although the number of training episodes was sufficient for observing convergence, it may not fully reflect long‑term stability or performance in more complex scenarios. Additional replicates or experiments with varied fire geometries would help assess how well these behaviors generalize. While the evaluation of metrics—distance to fire, fires viewed, episode length, and collisions—captured core aspects of performance, incorporating measures such as energy use or coverage uniformity could offer a more complete assessment. 

These findings should be interpreted with caution. Improved navigation and fire‑tracking in simulation does not guarantee similar performance in real wildfire conditions. The simulated environment omits critical real‑world factors such as wind, uneven terrain, sensor noise, and communication delays, all of which could influence drone behavior. Moreover, the observed relationships between training components and performance improvements do not establish direct causation; no single element—reward shaping, curriculum design, or multi‑agent coordination—can be identified solely responsible. Instead, the results suggest that these components collectively support the learning process. 

Future studies could explore more realistic fire dynamics such as the incorporation of environmental disturbances, for example wind, or communication constraints. Testing alternative reward structures or comparing different reinforcement learning algorithms could also help determine which approaches are most effective for wildfire monitoring tasks. Expanding the number of scenarios and increasing the diversity of fire shapes would further strengthen the generalizability of the findings. 

In summary, the results indicate that the trained agents developed safer and more structured navigation behaviors over the course of training. While these findings highlight the potential of reinforcement learning for wildfire monitoring, additional experiments and more complex simulations are needed to fully understand the strengths and limitations of this approach. The study provides a foundation for future work aimed at improving autonomous wildfire‑tracking systems and exploring broader applications in environmental monitoring. Together, these findings demonstrate the promise of DRL-driven UAV systems while highlighting the need for continued refinement in realistic, high-variability wildfire environments.

\section{Appendix}

\printbibliography{}
\end{twocolumn}
\end{document}